\def\thetitle{
Comprehensive Process Drift Detection\\with Visual Analytics%
}
\def\theshorttitle{Comprehensive Process Drift Detection with Visual Analytics}
\def\thekeywords{Process mining \and Process drifts \and Declarative process models}
\def\theauthors{
	Anton~Yeshchenko\inst{1}\orcidID{0000-0002-5346-8358}%
	\and %
	Claudio~Di~Ciccio\inst{1}\orcidID{0000-0001-5570-0475}%
	\and %
	Jan~Mendling\inst{1}\orcidID{0000-0002-7260-524X}%
	\and %
	Artem~Polyvyanyy\inst{2}\orcidID{0000-0002-7672-1643}%
}
\def\theauthorsshortlist{
	A.\ Yeshchenko, C.\ Di Ciccio, J.\ Mendling, A.\ Polyvyanyy
}
\def\theaffiliations{%
	Vienna University of Economics and Business, Vienna, Austria\\
	\email{\{anton.yeshchenko,claudio.di.ciccio,jan.mendling\}@wu.ac.at}
	\and
	The University of Melbourne, Parkville, VIC, 3010,
	Australia\\
	\email{artem.polyvyanyy@unimelb.edu.au}
}
\PassOptionsToPackage{table}{xcolor}
\documentclass[runningheads]{llncs}
\usepackage[textsize=scriptsize,backgroundcolor=yellow!40,obeyDraft]{todonotes} 
\usepackage[english]{babel}
\usepackage{subcaption}
\usepackage{url}
\usepackage{colortbl}
\usepackage{amssymb}
\usepackage{stmaryrd}
\usepackage{amsmath}
\usepackage{mathrsfs}
\usepackage{amssymb}
\usepackage[left]{lineno}
\usepackage[nonumberlist,acronym,sanitize=none]{glossaries}
\glsdisablehyper
\usepackage{comment}
\usepackage[pdftex, colorlinks=true, hyperfootnotes=true, hyperindex=true,
            plainpages=false, pagebackref=false, pdfpagelabels=true, pdfstartview=FitH,
            linkcolor=blue, citecolor=blue, urlcolor=blue,
            bookmarks, bookmarksopen, bookmarksdepth=3]{hyperref}
\usepackage[capitalise,nameinlink]{cleveref}
\Crefname{algocf}{Algorithm}{Algorithms}
\usepackage{tkz-base}
\usetikzlibrary{decorations.pathmorphing,trees,snakes,arrows,shapes,automata}
\usepackage{paralist}
\usepackage{multicol}
\usepackage{booktabs}
\usepackage{enumitem}
\usepackage{multirow}
\usepackage{rotating}
\usepackage{etaremune}
\usepackage{marginnote}
\usepackage{mathtools}
\usepackage{mathabx}
\usepackage{ifthen}
\usepackage[normalem]{ulem}
\usepackage{lineno}
\usepackage{soul}
\usepackage{xfrac}
\usepackage{floatrow}
\usepackage{lipsum}
\usepackage{makecell}
\usepackage{siunitx}
\usepackage{adjustbox}
\usepackage{microtype}
\usepackage{wrapfig}
\usetikzlibrary{DECLARE} 
\usepackage[hang,flushmargin,multiple]{footmisc}

\renewcommand{\arraystretch}{1.5}
\renewcommand\tabcolsep{5pt}
\newcolumntype{d}{>{\columncolor{gray!10}}c}
\newcolumntype{m}{>{\columncolor{gray!10}}l}
\newfloatcommand{capbtabbox}{table}[][\FBwidth]
{\begin{inparaenum}[\itshape(i)\upshape]}%
{\end{inparaenum}}

\def\GoodExampleMark{\checkmark}
\def\BadExampleMark{$\times$}

\RequirePackage{xparse}
\NewDocumentEnvironment{AuthNote}{+o+o}{%
	\IfValueT{#2}{\marginnote{\scriptsize{#2}}}%
	\begin{scriptsize}
		\colorbox{gray}%
		{\color{white} Note\IfValueT{#1}{ (#1)}:}%
		\quad%
		\color{brown}
}{%
	\normalcolor
	\end{scriptsize}
}

\RequirePackage{pifont}

\RequirePackage{lipsum}
\newcommand{\LipsumGray}[1][]{{\color{gray}\ifthenelse{\equal{#1}{}}{\lipsum}{\lipsum[#1]}}}

\RequirePackage{siunitx}
\newcolumntype{D}[1]{S[
	table-omit-exponent,
	round-mode=places,
	round-integer-to-decimal,
	round-precision={#1}]} 
\input{addon/macros-for-requirements-list}
\input{addon/macros-DECLARE} 
\input{addon/macros-DECLARE-drawing} 
\graphicspath{{figures/}}
\usepackage{times}

\usepackage[subtle]{savetrees}
\let\llncssubparagraph\subparagraph
\let\subparagraph\paragraph
\usepackage{titlesec}
\let\subparagraph\llncssubparagraph
\titlespacing{\subsubsection}{0pt}{*0.5}{*0.5}

\newacronym[\glslongpluralkey={Business Processes}]{bp}{BP}{Business Process}
\newacronym{wf}{WF}{workflow}
\newacronym{bpi}{BPI}{Business Process Intelligence}
\newacronym{bpm}{BPM}{Business Process Management}
\newacronym{bpms}{BPMS}{Business Process Management System}
\newacronym{bpmn}{BPMN}{Business Process Model and Notation}
\newacronym{cpn}{CPN}{colored Petri net}
\newacronym{soa}{SOA}{Service-Oriented Architecture}
\newacronym{kpi}{KPI}{Key Performance Indicator}
\newacronym{wfms}{WfMS}{Workflow Management System}
\newacronym{pn}{PN}{Petri net}
\newacronym{xes}{XES}{eXtensible Event Stream}
\newacronym{yawl}{YAWL}{Yet Another Workflow Language}
\newglossaryentry{task}{%
	name={task},description={the non-divisible, elementary activity}}
\def\paramx {\ensuremath{x}}
\def\paramy {\ensuremath{y}}

\def\letterx {\ensuremath{a}}
\def\lettery {\ensuremath{b}}
\def\letterz {\ensuremath{c}}

\newcommand{\Task}[1] {\ensuremath{\scalebox{0.85}{\normalfont\textsf{#1}}}}
\def\taska {\Task{a}}
\def\taskb {\Task{b}}
\def\taskc {\Task{c}}

\newglossaryentry{promod}{%
	name={process model},description={the model of a process}
}
\def\LogAlph {\ensuremath{\Sigma}}
\newglossaryentry{logalph}{
	name={log alphabet},description={the process alphabet, as reflected in a log},%
	symbol={\LogAlph}}
\def\Evt {\ensuremath{e}}
\newglossaryentry{evt}{
	name={event},description={a record of an instantaneous fact during the process enactment},%
	symbol={\Evt}}
\def\Trc { \ensuremath{\tau} }
\def\StrTrc { \ensuremath{\sigma} }
\newglossaryentry{trace}{
	name={trace},description={a sequence of \glsplural{evt}},%
	symbol={\Trc}}
\def\EvtLog {\ensuremath{L}}

\newglossaryentry{evtlog}{
	name={event log},description={a collection of \glstext{evttrace}s},%
	symbol={\EvtLog}} 
\newcommand{\DclrSty}[1] {\textsc{#1}}
\def\Declare {\DclrSty{Declare}}
\newglossaryentry{declare}{%
	name={\Declare},description={a declarative process modelling language and notation}}
\def\DeclaModel {\ensuremath{\mathcal{M}}}
\newglossaryentry{declamodel}{%
	name={declarative \glsentrytext{promod}},description={\glsentrydesc{promod}, expressed by means of constraints},
	symbol={\DeclaModel}
}

\newglossaryentry{mindeclamodel}{%
	name={discovered \glsentrytext{declamodel}},description={\glsentrydesc{declamodel}, discovered from an \glsentrytext{evtlog}},
	symbol={\DeclaModel}
}
\newglossaryentry{minerful}{%
	name={\DclrSty{miner}ful},description={A declarative process discovery algorithm}}
\newacronym{mf}{Mf}{\gls{minerful}}
\newglossaryentry{minerfulVac}{%
	name={MINERful Vacuity Checker},description={\glsentrytext{minerful}} algorithm with semantical vacuity detection}
\newacronym{mfv}{Mf-Vchk}{\gls{minerfulVac}}
\newacronym{dmm}{DMM}{Declare Maps Miner}
\newglossaryentry{decmapmin}{%
	name={Declare Maps Miner},description={the declarative process discovery algorithm \glsentrytext{decmapmin}}}
\newacronym{dmm2}{DM2}{Declare Miner 2}
\newglossaryentry{decmapmin2}{%
	name={Declare Miner 2},description={improvement of \glsentrytext{decmapmin} algorithm}}
\newglossaryentry{janus}{%
	name={Janus},description={the declarative process discovery algorithm \glsentrytext{janus}}}
\def\Subsum {\ensuremath{\sqsubseteq}}

\newglossaryentry{subsum}{%
	name={subsumption},description={is subsumed by},%
	symbol={\Subsum}}

\newglossaryentry{relaxop}{%
	name={relaxation},description={relaxation operator, climbing the \glsentrytext{subsum} hierarchy}}

\newglossaryentry{actv}{%
	name={activation},description={the activation of a constraint}}
\newglossaryentry{activator}{name={activator},description={the event that signals the occurrence of the activation in the trace}}

\newglossaryentry{target}{%
	name={target},description={target}}

\def\Cns {\ensuremath{C}}
\newglossaryentry{con}{%
	name={constraint},description={a temporal business rule},
	symbol={\Cns}
}

\newglossaryentry{welldef}{%
	name={well-defined},description={of \glsentrytext{con}s for which a finite non-empty trace exists that complies with them}
}
\newglossaryentry{cnspar}{%
	name={parameter},description={a parameter of a \glsentrytext{con}},
}
\newglossaryentry{cnsarity}{%
	name={arity},description={number of parameters of a \glsentrytext{con}},
}
\newglossaryentry{exi}{
	name={existence},
	description={constrains single tasks}
}
\newglossaryentry{exicon}{
	name={\glsentrytext{exi} \glsentrytext{con}},
	description={constrains single tasks}
}
\newglossaryentry{posicon}{
	name={position \glsentrytext{con}},
	description={constrains the position of tasks}
}
\newglossaryentry{cardicon}{
	name={cardinality \glsentrytext{con}},
	description={limits the number of tasks}
}
\newglossaryentry{rela}{
	name={relation},
	description={constraint on pairs of tasks}
}
\newglossaryentry{relacon}{
	name={\glsentrytext{rela} \glsentrytext{con}},
	description={constraint on pairs of tasks}
}
\newglossaryentry{unirelacon}{
	name={unidirectional \glsentrytext{relacon}},
	description={constraint on pairs of tasks, out of which one is the activation, as the other is the target}
}
\newglossaryentry{unifwrelacon}{
	name={\glsentrytext{fw}-\glsentrytext{unirelacon}},
	description={constraint on pairs of tasks, having the first parameter as the activation, and the second one as the target}
}
\def\FwCns {\ensuremath{\mathit{fw}}}
\newglossaryentry{fw}{
	name={forward},
	description={forward constraint},
	symbol={\FwCns}
}

\newglossaryentry{unibwrelacon}{
	name={\glsentrytext{bw}-\glsentrytext{unirelacon}},
	description={constraint on pairs of tasks, having the second parameter as the activation, and the first one as the target}
}
\def\BwCns {\ensuremath{\mathit{bw}}}
\newglossaryentry{bw}{
	name={backward},
	description={backward constraint},
	symbol={\BwCns}
}

\newglossaryentry{corelacon}{
	name={coupling \glsentrytext{con}},
	description={constraint based on pairs of relation constraints}
}
\newglossaryentry{nega}{
	name={negative},
	description={of a constraint, that negates a coupling relation constraint}
}
\newglossaryentry{negacon}{
	name={\glsentrytext{nega} \glsentrytext{con}},
	description={constraint negating a coupling relation constraint}
}
\def\CnsTmp {\ensuremath{\mathcal{C}}}
\newglossaryentry{cnstemp}{%
	name={template},description={the template of a \glsentrydesc{con}},
	symbol={\CnsTmp}}
\def\CnsTmpPrm {\ensuremath{\CnsTemp'}}
\def\CnsTmpSec {\ensuremath{\CnsTemp''}}
\newcommand{\CnsTmpFunc}[2] {\ensuremath{\CnsTmp(#1\ifthenelse{\equal{#2}{}}{}{,#2})}}
\newcommand{\CnsTmpFuncPrm}[2] {\ensuremath{\CnsTmpPrm(#1\ifthenelse{\equal{#2}{}}{}{,#2})}}
\newcommand{\CnsTmpFuncSec}[2] {\ensuremath{\CnsTmpSec(#1\ifthenelse{\equal{#2}{}}{}{,#2})}}

\newglossaryentry{cnstype}{%
	name={type},description={the type of a \glsentrydesc{cnstemp}}}
\def\CnsRep {\ensuremath{\mathfrak{C}}}
\newglossaryentry{cnsrep}{name={repertoire},description={the repertoire of \glsentrytext{declare} \glsentrytext{temp}s},
	symbol={\CnsRep}}
\newglossaryentry{cnsuniv}{name={\glsentrytext{con}s universe},description={the set of \glsentrytext{declare} \glsentrytext{temp}s over the process alphabet reflected in the log}}
\def\CnsInstRelation {\ensuremath{\Gamma}}

\newglossaryentry{cnsinst}{%
	name={\glsentrytext{cnstemp} instantiation relation},description={the assignment relation instantiating \glsentrytext{cnstemp}s into \glsentrytext{con}s, namely assigning \glsentrytext{task}s to \glsentrytext{cnspar}s.},
	symbol={\CnsInstRelation}}

\newcommand{\CnsInterpFun} {\ensuremath{\mathscr{I}}}
\newglossaryentry{cnsinterp}{
	name={interpretation function},description={function interpreting a \glsentrytext{declamodel}},
	symbol={\CnsInterpFun}}

\def\RelaConTemp {\ensuremath{\mathcal{R}}}
\newglossaryentry{relacontemp}{%
	name={relation template},description={the template of a relation \glsentrydesc{con}},
	symbol={\RelaConTemp}}

\def\ExiConTemp {\ensuremath{\mathcal{E}}}
\newglossaryentry{exicontemp}{%
	name={existence template},description={the template of an existence \glsentrydesc{con}},
		symbol={\ExiConTemp}}

\def\Supp {\ensuremath{\sigma}}

\newglossaryentry{support}{%
	name={support},description={the support of a \glsentrydesc{con}},
	symbol={\Supp}}

\def\Conf {\ensuremath{\kappa}}
\newglossaryentry{conf}{%
	name={confidence},description={the confidence level of a \glsentrydesc{con}},
	symbol={\Conf}}

\def\IntF {\ensuremath{\iota}}
\newglossaryentry{intf}{%
	name={interest factor},description={the interest factor of a \glsentrydesc{con}},
	symbol={\IntF}}

\def\CnsEvalFunctor {\ensuremath{\eta}}
\newglossaryentry{cnseval}{
	name={evaluation},description={evaluation of a \glsentrytext{con} or a \glsentrytext{declamodel} over a \glsentrytext{evttrace} or an \glsentrytext{evtlog}},
	symbol={\CnsEvalFunctor}}

\def\UniqTxt {AtMostOne}

\def\RespTxt {Response}
\def\AltRespTxt {AlternateResponse}

\def\ChaRespTxt {ChainResponse}
\def\PrecTxt {Precedence}
\def\AltPrecTxt {AlternatePrecedence}

\def\ChaPrecTxt {ChainPrecedence}

\def\NotSuccTxt {NotSuccession}

\def\UniqTmp {\ensuremath{\DclrSty{\UniqTxt}}}

\def\RespTmp {\ensuremath{\DclrSty{\RespTxt}}}
\def\AltRespTmp {\ensuremath{\DclrSty{\AltRespTxt}}}
\def\ChaRespTmp {\ensuremath{\DclrSty{\ChaRespTxt}}}

\def\AltPrecTmp {\ensuremath{\DclrSty{\AltPrecTxt}}}
\def\ChaPrecTmp {\ensuremath{\DclrSty{\ChaPrecTxt}}}

\def\NotSuccTmp {\ensuremath{\DclrSty{\NotSuccTxt}}}

\newcommand{\Uniq}[1] {\ensuremath{\DclrSty{\UniqTxt}(#1)}}

\newcommand{\Resp}[2] {\ensuremath{\DclrSty{\RespTxt}(#1,#2)}}

\newcommand{\AltResp}[2] {\ensuremath{\DclrSty{\AltRespTxt}(#1,#2)}}

\newcommand{\ChaResp}[2] {\ensuremath{\DclrSty{\ChaRespTxt}(#1,#2)}}

\newcommand{\Prec}[2] {\ensuremath{{\DclrSty{\PrecTxt}}(#1,#2)}}
\newcommand{\AltPrec}[2] {\ensuremath{\DclrSty{\AltPrecTxt}(#1,#2)}}

\newcommand{\ChaPrec}[2] {\ensuremath{\DclrSty{\ChaPrecTxt}(#1,#2)}}

\newcommand{\NotSucc}[2] {\ensuremath{\DclrSty{\NotSuccTxt}(#1,#2)}}

\newglossaryentry{fulfilment}{name={fulfilment},description={satisfaction of a constraint on a trace in which the activation occurs}} 
\newglossaryentry{condridet}{
	name={process drift detection},description={automated identification of changes in the process execution},%
}
\newacronym{dvd}{VDD}{Visual Drift Detection}
\def\DriftMap {Drift Map}
\def\DriftChart {Drift Chart}

\def\Windo{\ensuremath{\mathrm{win}}}
\def\WinSize{\ensuremath{\Windo_{\mathrm{size}}}} 
\def\WinStep{\ensuremath{\Windo_{\mathrm{step}}}} 
\def\WinNum{\ensuremath{\#_{\Windo}}} 
\def\CnsNum{\ensuremath{\#_{\mathrm{cns}}}} 
\def\Errtcsm{\ensuremath{\mathrm{Ertc}}}

\begin{document}
\title{\thetitle}
\titlerunning{\theshorttitle}
%
\author{\theauthors}
\authorrunning{\theauthorsshortlist}
%
\institute{\theaffiliations}
\maketitle              
%
%
%
 %
%
%
\begin{abstract}
Recent research has introduced ideas from concept drift into process mining to enable the analysis of changes in business processes over time. This stream of research, however, has not yet addressed the challenges of drift categorization, drilling-down, and quantification. In this paper, we propose a novel technique for managing process drifts, called~\gls{dvd}, which fulfills these requirements. The technique starts by clustering declarative process constraints discovered from recorded logs of executed business processes based on their similarity and then applies change point detection on the identified clusters to detect drifts. \gls{dvd} complements these features with detailed visualizations and explanations of drifts. Our evaluation, both on synthetic and real-world logs, demonstrates all the aforementioned capabilities of the technique.

\end{abstract}
\setcounter{footnote}{0}
%
%
%
%
\section{Introduction}
\label{sec:intro}
%
%

The availability of data has extended conceptual modeling as a research field of manually created models with automatic techniques for generating models from data. Process mining is one of these recent extensions that is concerned with providing transparency of how the businesses operate based on real-world event data. Process discovery algorithms have proven to be highly effective in generating process models from data of stable behavior~\cite{DBLP:books/sp/Aalst16}. However, many processes are not stable but are subject to various forms of change over time. In data mining, such change over time is called a \emph{drift}. A drift is a concept that process mining has addressed only to a limited extent so far.

Recent works have focused on integrating ideas from research on concept drift from data mining into process mining~\cite{Denisov/BPM2018:MiningConceptDriftinPerformanceSpectraofProcesses,DBLP:conf/simpda/HompesBADB15,DBLP:conf/s-bpm-one/SeeligerNM17,DBLP:conf/otm/ZhengW017,DBLP:conf/er/OstovarMRHD16}. The arguably most advanced technique is proposed in~\cite{DBLP:journals/tkde/MaaradjiDRO17}, where Maaradji et al. present a framework for detecting process drifts based on tracking behavioral relations over time using statistical tests. A strength of this approach is its statistical soundness and ability to identify a rich set of drifts, which makes it a suitable tool for verifying if an intervention at a known point in time has resulted in an assumed change of behavior. However, in practice, the existence of different types of drifts in a business process is not known beforehand, and the analysts are interested in distinguishing what has and what has not changed over time. This need calls for a more fine-granular analysis.

In this paper, we present a novel technique for \gls{condridet}, called \acrfull{dvd}, which addresses the identified research gap. More specifically, our technique facilitates the \textit{visual interpretation}~\cite{ware2012information} of process drifts founded in the formal rigor of temporal logic of {\Declare} constraints~\cite{Aalst.etal/CSRD09:DeclarativeWFsBalancing,DBLP:journals/tmis/CiccioM15} and time series analysis~\cite{CharlesTruonga/SParxiv:SelectiveReviewOfOfflineChangePointDetectionMethods}. Key strengths of our technique are clustering, i.e., grouping, of declarative behavioral constraints that exhibit similar trends of changes over time and automatic detection of changes, i.e., drift points. These features allow us to detect and explain drifts that would otherwise sneak undetected by other techniques. The paper presents an evaluation that demostrates these capabilities.

The remainder of the paper is structured as follows. \Cref{sec:background} illustrates the problem of process drift detection and formulates five requirements for its analysis. Then, \Cref{sec:preliminaries} states the preliminaries. \Cref{sec:approach} presents our drift detection technique, while \Cref{sec:evaluation} evaluates the technique using synthetic and real-world benchmark data. Finally, \Cref{sec:conclusion} summarizes the results and concludes with an outlook on future research.

%
%
\section{Process Drift Analysis}
\label{sec:background}
%
%

This section discusses and motivates the problem of process drift analysis (\cref{sec:drifts}), and specifies requirements for its solution (\cref{sec:problem}). 

\subsection{Motivating example}
\label{sec:drifts}
%
%

Various logs of real-world business process executions have been recently made available for research. 
As an example, consider the log of the Italian process for handling the collection of road ticket fines~\cite{DBLP:journals/computing/MannhardtLRA16}.
This process starts with a ticket being issued. 
In the best case, which covers a third of all the cases, the fine is directly paid. 
In roughly half of the other cases, a fine notification is sent to the accused driver. 
Some of these drivers appeal, while some ignore the notice, such that a considerable share of cases sees a penalty being added. 
Partially, these are further appealed, paid or eventually sent for credit collection. The authority is now interested in this question: 
Has the process of handling road ticket fines, specifically for the accused drivers, changed over time, and which parts of the process now work differently than in the past?

The described problem is typical for many domains. 
The objective is to explain the change of the system's behavior in a dynamically changing non-stationary environment based on some \textit{hidden context}~\cite{DBLP:journals/csur/GamaZBPB14}. 
In this setting, a \textit{concept drift} is a change of the conditional distribution of the output given a specific input. 
Research in data mining and machine learning distinguishes techniques for uncovering drifts in an \textit{online} or \textit{offline} manner~\cite{TsymbalAlexey/:TheProblemOfConceptDrift:DefinitionsAndRelatedWork}, with applications in prediction and fraud detection.

\begin{figure}[t]
	\includegraphics[width=\columnwidth]{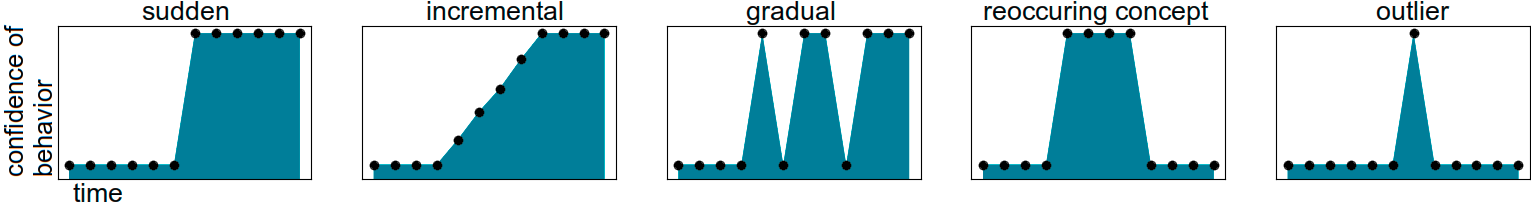}
	\vspace{-3mm}
	\caption{Different types of drifts, cf.\ Fig.\ 2 in~\cite{DBLP:journals/csur/GamaZBPB14}; note that an outlier is not a drift.}
	\label{fig:driftypes2}
	\vspace{-3mm}
\end{figure}

In process mining, \emph{process drift} is a notion for analyzing changes of business processes over time. 
Classical process mining techniques have implicitly assumed that logs are not sensitive to time in terms of systematic change~\cite{DBLP:books/sp/Aalst16}. 
Sampling-based techniques explicitly build on this assumption for generating a process model with a subset of the event log data~\cite{DBLP:conf/caise/BauerSGGW18}. 
A significant challenge for adopting concept drift for process mining is to represent behavior in a time-dependent way. 
The approach reported in~\cite{DBLP:journals/tkde/MaaradjiDRO17} uses causal dependencies and tracks them over time windows. 
The specific challenge is to not only spot a drift but also to classify it. 
\Cref{fig:driftypes2} shows established drift classes from data mining. 
Next, we use the example of the road ticket fines process to illustrate the potential causes of drifts. 

A \textit{sudden drift} is typically caused by an intervention. 
A new law could eliminate the right of an accused driver to lodge a second appeal. As a result, we would not see second appeal events in our log in the future. 
An \textit{incremental drift} might result from a stepwise introduction of self-service terminals for paying fines at toll stations. 
A \textit{gradual drift} may yield from a new policy to show less indulgence with drivers who marginally violated speeding rules.
Finally, a \textit{reoccurring drift} might result from specific measures taken in the holiday season from June to August, like flagging down drivers directly on the highway to have them pay right on the spot.
Existing process mining techniques support these types of drifts partially.

\begin{sloppypar}
The following are four cases from the Italian road ticket fines log\footnote{\scriptsize \url{https://doi.org/10.1007/s00607-015-0441-1}}: 
\begin{compactenum}
	\item 10 Jan.\ 2011: $\langle \Task{Lodging ticket},\; \Task{Appeal},\; \Task{Appeal},\; \Task{Payment},\; \Task{Close ticket} \rangle$
	\item 15 Jan.\ 2011: $\langle \Task{Lodging ticket},\; \Task{Appeal},\; \Task{Appeal},\; \Task{No payment},\; \Task{Close ticket} \rangle$
	\item 04 Feb.\ 2011: $\langle \Task{Lodging ticket},\; \Task{Appeal},\; \Task{Payment},\; \Task{Close ticket} \rangle$
	\item 06 Feb.\ 2011: $\langle \Task{Lodging ticket},\; \Task{Appeal},\; \Task{No payment},\; \Task{Close ticket} \rangle$
\end{compactenum}
We observe a sudden drift here due to the introduction of a new law. After 4 Feb. 2011, it is not possible to lodge a second appeal. Therefore, in formal terms, from case 3 onwards,  
the behavioral rule that multiple appeals occur before the ticket closes abruptly decreases in confidence. In {\Declare}, we denote this rule as $\AltResp{\Task{Appeal}}{\Task{Close ticket}}$.

\end{sloppypar}

\subsection{Requirements}
\label{sec:problem}
%
%

Based on the analysis of process change scenarios from the literature, like the road ticket fines discussed previously, we identified five requirements for process drift analysis:

%
\begin{requidescr} 
	\item[Identify drifts:\namedlabel{req:identify}] The points at which a business process undergoes drifts should be identified based on precise criteria; 
	\item[Categorize drifts:\namedlabel{req:categorize}] Process drifts should be according to their types; 
	\item[Drill down and roll up analysis:\namedlabel{req:drill}] Process drifts should be characterized at different levels of granularity, e.g., drifts that concern the entire process or only its parts; 
	\item[Quantitative analysis:\namedlabel{req:quanti}] Process drifts should be associated with a degree of change, a measure that quantifies to which extent the drift entails a change in the process;
	\item[Qualitative analysis:\namedlabel{req:quali}] Process drifts should convey changes in a business process to process analysts effectively.
\end{requidescr}
\setlength\intextsep{0pt}
\begin{wraptable}[12]{r}{.6\textwidth}
	\vspace{-2mm}
	\caption{Process drift detection in process mining.}
	\label{table-approaches}
\begin{tabular}{l r r r r r}
\toprule
\textbf{Approach} & \textbf{R1} & \textbf{R2}& \textbf{R3} & \textbf{R4} & \textbf{R5} \\

\midrule
ProDrift~\cite{DBLP:journals/tkde/MaaradjiDRO17,DBLP:conf/er/OstovarMRHD16} & + & +/- & - & - & - \\
TPCDD~\cite{DBLP:conf/otm/ZhengW017} & + & - & - & - & - \\
Process Trees~\cite{quteprints121158} & + & - & - & - & + \\
Performance Spectra~\cite{Denisov/BPM2018:MiningConceptDriftinPerformanceSpectraofProcesses} & - & - & + & - & + \\
Comparative Trc.\ Clustering~\cite{DBLP:conf/simpda/HompesBADB15} & - & - & - & + & + \\
Graph Metrics On Proc.Graphs~\cite{DBLP:conf/s-bpm-one/SeeligerNM17} & + & - & - & + & + \\ \midrule
\textbf{\Gls{dvd} approach (this paper)}  & \textbf{+} & \textbf{+} & \textbf{+} & \textbf{+} & \textbf{+} \\
\bottomrule
\end{tabular}
%
\end{wraptable}

\noindent

\Cref{table-approaches} provides an overview of the state-of-the-art methods to process drift analysis with the reference to the requirements. 
Note that collectively these methods implement all the requirements, whereas each individual methods addresses only a subset thereof.

Approaches like ProDrift~\cite{DBLP:journals/tkde/MaaradjiDRO17} and Graph Metrics on Process Graphs~\cite{DBLP:conf/s-bpm-one/SeeligerNM17} put an emphasis on requirement \ref{req:identify}. 
The evaluation of ProDrift in~\cite{DBLP:journals/tkde/MaaradjiDRO17} shows that two types of drifts are found with high accuracy (sudden and gradual drifts), hence partly addressing requirement~\ref{req:categorize}; note that the authors report high sensitivity of the technique to the choice of the method parameters. The approach relies on the automated detection of changes in business process executions, which are analyzed based on causal dependency relations studied in process mining~\cite{1316839}. The Tsinghua Process Concept Drift Detection approach (TPCDD)~\cite{DBLP:conf/otm/ZhengW017} uses two kinds of behavioral relationships (direct succession and weak order). The approach computes those relations on every trace, so as to later identify the change points through their merge and clustering. The sole type of drift that TPCDD detects is the sudden drift.

The other approaches emphasize requirement~\ref{req:quali}.
The approach based on Process Trees~\cite{quteprints121158} uses ProDrift for drift detection, and aims at explaining how sudden drifts influence behavior of the process. To this end, process trees for pre-drift and post-drift sections of the log are built and used to explain the change. The Performance Spectra approach~\cite{Denisov/BPM2018:MiningConceptDriftinPerformanceSpectraofProcesses}
focuses on drifts that show seasonality. The technique filters the control-flow and visualizes identified flow patterns. It is evaluated against a real-world log, in which recorded business processes show year-to-year seasonality. 
A strength of the Comparative Trace Clustering approach~\cite{DBLP:conf/simpda/HompesBADB15} is its ability to include non-control-flow characteristics in the analysis. 
Based on these characteristics, it partitions and clusters the log. The differences between the clusters, then, indicate the quantitative change in the business processes, refer to requirement~\ref{req:quanti}.
The Graph Metrics on Process Graphs approach~\cite{DBLP:conf/s-bpm-one/SeeligerNM17} discovers a first model, called a reference, using the Heuristic Miner on a section of the log~\cite{DBLP:books/sp/Aalst16}. Then, it discovers models for other sections of the log and uses graph metrics to compare them with the reference model. 
The technique interprets significant differences in the metrics as drifts.
The reference model and detection windows get updated, once a drift is detected.

This discussion, summarized in \cref{table-approaches}, witnesses that none of the state-of-the-art methods addresses all the five requirements. 
Thus, the work at hand, to address the gap.

%
%
\section{Preliminaries}
\label{sec:preliminaries}
%
%
In this section, formal preliminaries of the approach are given. \Cref{sec:declare} discusses {\Declare} specification as the main body of process mining research we build upon. \Cref{sec:clusteringChangepoint} describes clustering and change point detection methods, which are the main instruments of our approach.

An event log $L$ (\emph{log} for short) is a collection of recorded traces that correspond to process executions.
In this paper, we abstract the set of activities of a process as a finite non-empty alphabet $\Sigma = \{ \letterx, \lettery, \letterz, \ldots \}$, and we define a trace as a finite sequence of activities $a_i \in \sigma, 1 \leq i \leq n$. 
Case 1 of the road ticket process from~\cref{sec:drifts} is an example of a trace. Cases 1-4 are an example of an event log. In the following examples, we shall also resort on the string-representation of traces (i.e., $\sigma=a_1 a_2 \cdots a_n$) defined over $\Sigma$.
Event log $L$ is a multiset of traces, as the same trace can be repeated multiple times in the same log: denoting the multiplicity $m \geqslant 0$ as an exponent of the trace, we have that
$L=\{ \sigma_1^{m_1}, \sigma_2^{m_2}, \ldots, \sigma_N^{m_N} \}$ (if $m_i=0$ for some $1 \leqslant i \leqslant N$ we shall simply omit $\sigma_i$).
The size of the log is defined as $|L| = \sum_{i=1}^{N}{m_i}$, i.e., the sum of its traces' multiplicities.
For example, the size of the Italian help desk log is \num{150370}.
A sub-log $L' \subseteq L$ of $L$ is a log $L'= \lbrace \sigma_1^{m'_1}, \sigma_2^{m'_2}, \ldots, \sigma_N^{m'_N} \rbrace$ such that $m'_i \leqslant m_i$ for all $1 \leqslant i \leqslant N$.
A log consisting of cases 1-3 from the example log $L$ in \cref{sec:drifts} is a sub-log of $L$.

\subsection{{\Declare} modeling and mining}
\label{sec:declare}
%
%
\begin{table}[tb]
	\caption{Example {\Declare} constraints.} 
	\label{tab:declare:verbose:gfx}
	\centering
	\resizebox{0.99\columnwidth}{!}{%
\renewcommand{\arraystretch}{1.6}
\rowcolors{2}{white}{gray!12.5}
\begin{tabular}{ l p{5cm} l l l l }
	\toprule
	\textbf{Constraint} & 
	\textbf{Explanation} & 
	\multicolumn{4}{c}{\textbf{Examples}}
	\\
	\midrule
%
%
%
	
	$\Uniq{\taska}$ & 
	If $\taska$ occurs, then it occurs at most \emph{once} &
	\GoodExampleMark \Task{bcc} & \GoodExampleMark \Task{bcac} &
	\BadExampleMark \Task{bcaac} & \BadExampleMark \Task{bcacaa}
	\\
	
%
%
	$\Resp{\taska}{\taskb}$ &
	If {\taska} occurs, then {\taskb} occurs eventually after {\taska} &
	\GoodExampleMark \Task{baabc} & \GoodExampleMark \Task{bcc} &
	\BadExampleMark \Task{caac} & \BadExampleMark \Task{bacc}
	\\
	$\AltResp{\taska}{\taskb}$ &
	If {\taska} occurs, then {\taskb} occurs eventually afterwards, and no other {\taska} recurs in between &
	\GoodExampleMark \Task{cacb} & \GoodExampleMark \Task{abcacb} &
	\BadExampleMark \Task{caacb} & \BadExampleMark \Task{bacacb}
	\\
	$\ChaResp{\taska}{\taskb}$ &
	If {\taska} occurs, then {\taskb} occurs immediately afterwards &
	\GoodExampleMark \Task{cabb} & \GoodExampleMark \Task{abcab} &
	\BadExampleMark \Task{cacb} & \BadExampleMark \Task{bca}
	\\
	$\Prec{\taska}{\taskb}$ &
	If {\taskb} occurs, then {\taska} must have occurred before &
	\GoodExampleMark \Task{cacbb} & \GoodExampleMark \Task{acc} &
	\BadExampleMark \Task{ccbb} & \BadExampleMark \Task{bacc}
	\\
	$\AltPrec{\taska}{\taskb}$ &
	If {\taskb} occurs, then {\taska} must have occurred before and no other {\taskb} recurs in between &
	\GoodExampleMark \Task{cacba} & \GoodExampleMark \Task{abcaacb} &
	\BadExampleMark \Task{cacbba} & \BadExampleMark \Task{abbabcb}
	\\
	$\ChaPrec{\taska}{\taskb}$ &
	If {\taskb} occurs, then {\taska} occurs immediately beforehand &
	\GoodExampleMark \Task{abca} & \GoodExampleMark \Task{abaabc} &
	\BadExampleMark \Task{bca} & \BadExampleMark \Task{baacb}
	\\
	$\NotSucc{\taska}{\taskb}$ &
	{\taska} occurs if and only if {\taskb} does not occur afterwards &
	\GoodExampleMark \Task{bbcaa} & \GoodExampleMark \Task{cbbca} &
	\BadExampleMark \Task{aacbb} & \BadExampleMark \Task{abb}
	\\
	\bottomrule
\end{tabular}%
	}
\end{table}
A declarative process specification represents the behavior of a process by means of \emph{constraints}, i.e., temporal rules that specify the conditions under which activities may, must, or cannot be executed.
In this paper we focus on {\Declare}, one of the most well-established declarative process modeling languages to date \cite{Aalst.etal/CSRD09:DeclarativeWFsBalancing}.

{\Declare} provides a standard library of templates (\emph{repertoire} \cite{Polyvyanyy.etal/FAOC2016:ExpressivePowerBehavioralProfiles,DiCiccio.etal/IS2017:ResolvingInconsistenciesRedundanciesDeclare}), i.e., constraints parametrized over activities.
Examples of {\Declare} constraints are
\Resp{\taska}{\taskb}
and
\ChaPrec{\taskb}{\taskc}.
The former constraint applies the {\RespTmp} template on tasks {\taska} and {\taskb}, 
and states that if {\taska} occurs then {\taskb} must occur later on within the same trace.
In this case, {\taska} is named \emph{activation}, because it is mentioned in the ``if'' clause, thus triggering the constraint, whereas {\taskb} is named \emph{target}, as it is in the consequent clause~\cite{DiCiccio.etal/IS2017:ResolvingInconsistenciesRedundanciesDeclare}.
\ChaPrec{\taskb}{\taskc} asserts that if {\taskc} (the activation) occurs, then {\taskb} (the target) must have occurred immediately before.
Given an alphabet of activities $\Sigma$, we denote the number of all possible constraints that derive from the application of {\Declare} templates to all activities in $\Sigma$ as $\CnsNum \subseteq O(\Sigma^2)$~\cite{DiCiccio.etal/IS2017:ResolvingInconsistenciesRedundanciesDeclare}.
For the Italian road ticket fine log, $\CnsNum = \num{1584}$.
\Cref{tab:declare:verbose:gfx} shows some of the templates of the {\Declare} repertoire, together with the examples of traces that satisfy (\GoodExampleMark) or violate (\BadExampleMark) them.

Declarative process mining tools can measure to what degree constraints hold true in a given event log~\cite{DBLP:conf/cidm/MaggiMA11}.
To that end, diverse measures have been introduced.
Among them, we consider here \emph{support} and \emph{confidence}~\cite{DBLP:journals/tmis/CiccioM15}.
Their values range from \num{0} to \num{1}.
In \cite{DBLP:journals/tmis/CiccioM15}, the support of a constraint is measured as the ratio of times that the event is triggered and satisfied over the number of activations.
Let us consider the following example event log:
$\EvtLog = \lbrace {\StrTrc_{1}^{4}}, {\StrTrc_{2}^{1}}, {\StrTrc_{3}^{2}} \rbrace$,
having
${\StrTrc_{1}} =
\Task{baabc}  $,
${\StrTrc_{2}} =
\Task{bcc} $, and
${\StrTrc_{3}} =
\Task{bcba} $.
The size of the log is $4+1+2=7$.
The activations of \Resp{\taska}{\taskb} that satisfy the constraint amount to \num{8} because two {\taska}'s occur in ${\StrTrc_{1}}$ that are eventually followed by an occurrence of {\taskb}, and ${\StrTrc_{1}}$ has multiplicity \num{4} in the event log. The total amount of the constraint's activations in $L$ is \num{10} (see the violating occurrence of {\taska} in ${\StrTrc_{3}}$). The support thus is \num{0.8}.
By the same line of reasoning, the support of \ChaPrec{\taskb}{\taskc} is $\frac{7}{8}=\num{0.875}$ (notice that in ${\StrTrc_{2}}$ only one of the two occurrences of {\taskc} satisfies the constraint).
To take into account the frequency with which constraints are triggered, confidence scales support by the ratio of traces in which the activation occurs at least once.
Therefore, the confidence of \Resp{\taska}{\taskb} is $\num{0.8} \times \frac{6}{7} \approx \num{0.69}$ because {\taska} does not occur in ${\StrTrc_{2}}$. As $\taskb$ occurs in all traces, the confidence of \ChaPrec{\taskb}{\taskc} is \num{0.875}.

\subsection{Clustering and change point detection algorithms}
\label{sec:clusteringChangepoint}
%
%
In this paper, we focus on the analysis of time-series data.
A \emph{time series} is a sequence of ordered data points $\left\langle t_1, t_2, \cdots, t_d\right\rangle = T \in \mathbb{R}^d$ consisting of
$d \in \mathbb{N^{+}}$ real values.
\Cref{fig:loop-detail} illustrates 
an example of time series.
A \emph{multivariate time series} is a set of $n \in \mathbb{N^{+}}$ time series $D=\{T_{1}, T_{2}, \ldots, T_{n}\}$.
We assume a multivariate time series to be piece-wise stationary except for its \emph{change points}.

\noindent
In our approach, we take advantage of the following techniques.

\subsubsection{Time series clustering}
\label{sub-sub-clustering}
is an unsupervised data mining technique for organizing data points into groups based on their similarity~\cite{Aghabozorgi:2015:TCD:2799194.2799230}.
The objective is to maximize data similarity within clusters and minimize it across clusters.
More specifically, the \emph{time-series clustering} is the process of partitioning $D$ into non-overlapping clusters of multivariate time series, $C=\{C_{1}, C_{2}, \ldots,C_{m}\} \subseteq 2^{D}$, with $C_{i} \subseteq D$ and $1\leq m \leq n$, for each $i$ such that $1 \leq i \leq m$, such that homogeneous time series are grouped together based on a \emph{similarity measure}.
A \emph{similarity measure} $\mathrm{sim}(T,T')$ represents the distance between two time series $T$ and $T'$ as a non-negative number. Time-series clustering is often used as a subroutine of other more complex algorithms and is employed as a standard tool in data science for anomaly detection, character recognition, pattern discovery, visualization of time series~\cite{Aghabozorgi:2015:TCD:2799194.2799230}.

\subsubsection{Change point detection}
\label{sub-sub-changepoint}
is a technique to detect the points in which multivariate time series exhibit changes in their values~\cite{CharlesTruonga/SParxiv:SelectiveReviewOfOfflineChangePointDetectionMethods}. 
Let $D^{j}$ denote all elements of $D$ at position $j$, i.e., $D^{j}=\{T_{1}^{j}, T_{2}^{j}, ..., T_{n}^{j}\}$, where $T^{j}$ is a $j$-th element of time series $T$.
The objective of change point detection algorithms is to find $k \in \mathbb{N^{+}}$ changes in $D$, where $k$ is previously unknown. Every element $D^{j}$ for $0 < j \leqslant k$ is a point at which the values of the time series undergo significant changes.
In \cref{fig:loop-detail}, e.g., each vertical black dashed line is one of the $k=9$ change points.
To detect change points, the search algorithms require a \emph{cost function} and a \emph{penalty} parameter as inputs.
The former describes how homogeneous the time series is. It is chosen in a way that its value is high if the time series contains many change points and low otherwise. The latter is needed to constrain the search depth. 
The supplied penalty should strike a good balance between finding too many change points and not finding any significant ones.
Change point detection is a technique commonly used in signal processing and, more in general, for the analysis of dynamic systems that are subject to changes~\cite{CharlesTruonga/SParxiv:SelectiveReviewOfOfflineChangePointDetectionMethods} . 

%
%
\section{Technique}
\label{sec:approach}
%
%
\begin{figure}[tb]
	\includegraphics[width=1.0\textwidth]{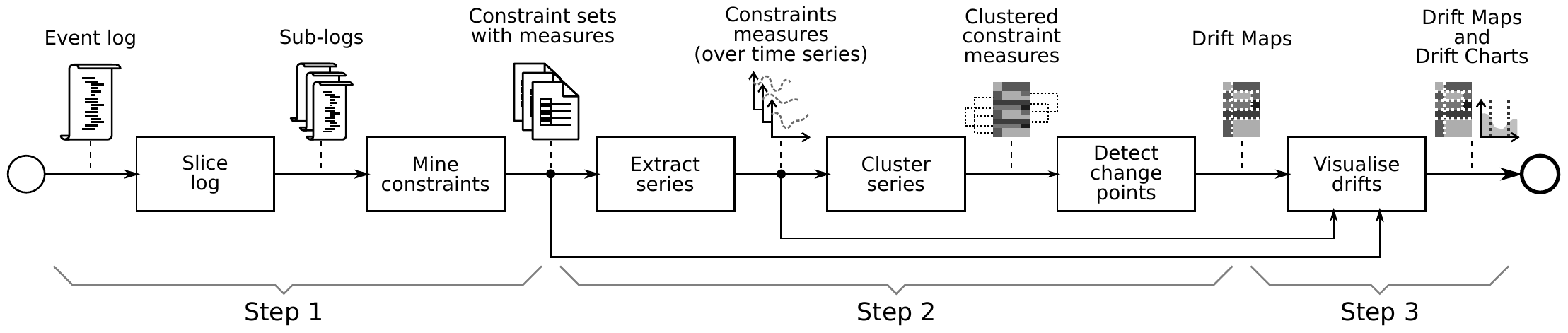}
	\caption{The \gls{dvd} approach.}
	\label{fig:approach}
\end{figure}
In this section, we introduce the \gls{dvd} approach.
First, we derive a multivariate time series from an event log, where each time series represents how the confidence values of some {\Declare} constraint evolve over time. 
We prefer confidence over support to prevent that sporadically occurring activities bias our detection algorithms. Then, we cluster sub-sets of time series to group together the constraints that expose a similar trend in their confidence value.
Next, using change point detection techniques, we identify the process drifts, i.e., the points in which significant changes in the confidence of behavioral rules occur.
Finally, we assess and explain behavioral changes through visual diagrams and numerical reports on drift metrics. \Cref{fig:approach} illustrates the multi-step \gls{dvd} approach.

\noindent
\textbf{Step 1: Mining {\Declare} windows.} In this step, we split the log into sub-logs. From each sub-log, we mine the set of {\Declare} constraints and compute their confidence.
\\
\textbf{Step 2: Slicing the {\Declare} constraints space into time and behavior sub-spaces}.
This step begins with the extraction of multi-variate time series that represent the trends of the constraints' confidence.
Thereupon, we cluster those time series to find groups of constraints that exhibit similar confidence trends (henceforth, \emph{behavior clusters}).
The step ends by returning the detected change points both in the entire multi-variate time series and in each cluster, so as to find overall and behavior-specific drifts, respectively.
\\
\textbf{Step 3: Explaining drifts.}
In the last step, we plot {\DriftMap}s and {\DriftChart}s to visually identify 
and characterize the detected drift. 
\noindent
In the following, we detail those steps.

\subsection{Mining {\Declare} windows}
\label{sec:declare:shift}

The first step takes as input a log $L$, and two additional parameters ({\WinSize} and {\WinStep}). It returns a multivariate time series $D$ based on the confidence of mined {\Declare} constraints. 

First, we sort the traces in the event log $L$ by the timestamp of their respective first events. Thereupon, we extract a sub-log from $L$ as a window of size $\WinSize \in \mathbb{N^{+}}$, with $1 \leqslant \WinSize \leqslant |L|$. We subsequently shift the sub-log window by a given step ($\WinStep \in \mathbb{N^{+}}$, with $1 \leqslant \WinStep \leqslant \WinSize$). Notice that we have sliding windows if $\WinStep < \WinSize$ and tumbling windows if $\WinStep = \WinSize$. Thus, the number of produced sub-logs is equal to:
$\WinNum = \left\lfloor \frac{|L| - \WinSize - \WinStep}{\WinStep} \right\rfloor$. Having {\WinSize} set to \num{5000} and {\WinStep} set to \num{2500}, {\WinNum} amounts to \num{57} on the Italian road fine ticket log.

For every sub-log $L_j \subseteq L$ thus formed ($1 \leqslant j \leqslant \WinNum$), we check all possible {\Declare} constraints that stem from the activities alphabet of the log, amounting to {\CnsNum} (see \cref{sec:declare}). 
For each constraint $i \in 1..\CnsNum$, we compute its confidence over the sub-log $L_j$, namely $\mathrm{Conf}_{i,j} \in [0,1]$.
This generates a time series $T_i =  ( \mathrm{Conf}_{i,1}, \ldots, \mathrm{Conf}_{i,\WinNum} ) \in [0,1]^{\WinNum}$ for every constraint $i$. In other words, every time series $T_i$ describes the confidence of all the {\Declare} constraints discovered in the $i$-th window of the event log. The multivariate time series $D=\{T_{1}, T_{2}, \ldots, T_{\CnsNum}\}$ encompasses the full spectrum of all constraints.
Next, we detail the steps of slicing the {\Declare} constraints and explaining the drifts.

\subsection{Slicing the {\Declare} constraints space into time and behavior sub-spaces}
\label{sec:cluster-changepoint}
The second step processes the previously generated multivariate time series $D$ to derive (i) a set $C$ of clusters exhibiting similar confidence trends, and (ii) a set of $k \in \mathbb{N^{+}}$ change points representing the process drifts.

\subsubsection{Change point detection.}
To detect change points, we use the \emph{Pruned Exact Linear Time (PELT)} algorithm~\cite{killick2012optimal}. This algorithm performs an exact search, but requires the input dataset to be of limited size. Our setup is appropriate as by design the length of the multivariate time-series is limited by the choice of parameters {\WinSize} and {\WinStep}. Also, this algorithm is suitable for cases in which the number of change points is unknown a priori~\cite[p.~24]{CharlesTruonga/SParxiv:SelectiveReviewOfOfflineChangePointDetectionMethods}, as in our case. We use the \emph{Kernel cost function}, detailed in~\cite{CharlesTruonga/SParxiv:SelectiveReviewOfOfflineChangePointDetectionMethods}, which is optimal for our technique,
and
adopt the procedures described in~\cite{killick2012optimal} to identify the optimal \emph{penalty} value. 

\subsubsection{Clustering time series of {\Declare} constraints.}
By applying a change point detection algorithm on the entire multivariate time-series, we are able to pinpoint the window (i.e., the sub-log) where overall behavior changes occur. However, the level of granularity may be inappropriate as we could not single out the phenomena that are local to certain behavioral rules. That would interfere with the accuracy of results. 
Therefore, we use time-series clustering techniques~\cite{Aghabozorgi:2015:TCD:2799194.2799230} to group together similarly changing pockets of behavior of the process. One time series describes how one constraint changes its confidence over time. By clustering, we find all the time series that share similar trends of values, hence, we find all similarly changing constraints. 
We use \emph{hierarchical clustering}, as it is reportedly one of the most suitable algorithms when the number of clusters is unknown~\cite{Aghabozorgi:2015:TCD:2799194.2799230}.
As a result, we obtain a partition of the multivariate time series of {\Declare} constraint confidence values into behavior clusters.

\subsection{Explaining drifts}
After clustering the behavior of the log and finding the change points, we expand the classification of the types of drifts found in the literature by being able to identify, pinpoint, and categorize the drifts within behavior clusters.
We also allow for an assessment of how erratic the clusters are by means of the novel measure described next.

\subsubsection{Finding erratic behavior clusters.}
\label{sec:erratic-measure}
The behavioral changes in one cluster can be visually depicted by a plot like that in~\cref{fig:loop-detail}. Thus, in order to find and pinpoint the most interesting (erratic) behavior clusters, we define a measure 
inspired by the idea of finding the length of a poly-line in a plot. The rationale is, straight lines denote a regular trend and have the shortest length, whilst more irregular, wavy curves evidence more behavior changes and their length is higher.
We are, therefore, mostly interested in longer lines.

We compute our measure as follows.
We calculate for all constraints $i$ such that $1 \leq i \leq \CnsNum$ the Euclidean distance $\delta : [0,1]\times[0,1]\to\mathbb{R_+}$ between consecutive values in the time series $T_i = (T_{i,1},\ldots,T_{i,\WinSize})$, i.e.,
$\delta(T_{i,j},T_{i,j+1})$ for every $j$ s.t.\ $1 \leqslant j \leqslant \WinSize$.
For every time series $T_i$, we thus derive the overall measure $\Delta(T_i)=\sum_{j=1}^{\WinSize-1}{\delta(T_{i,j},T_{i,j+1})}$.
Thereupon, to measure how erratic a behavior cluster is, we devise the following measure:
\begin{equation}
\label{eqn:erratic}
\Errtcsm(C) =  \sum_{j=1}^{|C|} \sqrt{1 + ( \Delta(T_i) \times \WinNum)^{2}}
\end{equation}
The most erratic behavior cluster has the highest $\Errtcsm$ value.

\subsubsection{Visual drift classification.}
\label{sec:visualdrift}
We enable the visual identification of the patterns illustrated in \cref{fig:driftypes2} with a graphical representation that we name {\DriftMap}s: they depict clusters and their constraints' confidence measure evolution along the time series, together with the drift points.
We allow the user to inspect every single cluster and its drifts in dedicated diagrams that we name {\DriftChart}s.

{\DriftMap}s, such as those illustrated in \cref{fig:conditionalMove} or \cref{fig:bpic2011:separate}, plot all drifts data on a two-dimensional plane. The visual representation we adopt is inspired by~\cite{ware2012information}. The x-axis is the time axis, while every constraint corresponds to a point along the y-axis. We add vertical lines to mark the identified change points, i.e., drift points, and horizontal lines to demark clusters. Constraints are sorted by the similarity of the confidence trends. The values of the time series are represented through the plasma color-blind friendly color map~\cite{ware2012information}, from blue (low peak) to yellow (high peak).

To analyze the time-dependent trend of specific clusters, we build {\DriftChart}s, such as those depicted in \cref{fig:loop-detail} or \cref{fig:bpic2011:erratic}. They have time on the x-axis and average confidence of the constraints in a cluster on the y-axis.
We add vertical lines as in {\DriftMap}s.

{\DriftMap}s permit the users to have a global picture of the clusters and of the process drifts. {\DriftChart}s allow for a visual categorization of the drifts according to the classification introduced in \cite{DBLP:journals/csur/GamaZBPB14} (\cref{fig:driftypes2}).
The following section demonstrates applications of this visual-aided approach on synthetic and real-world logs.

%
%
\section{Evaluation}
\label{sec:evaluation}
%
%

This section presents our evaluation setup, its results on detecting and explaining drifts, and a discussion of the results.

\subsection{Evaluation setup}

\setlength\intextsep{0pt}
\begin{wraptable}[9]{r}{.5\textwidth}
\vspace{-2mm}
\caption{Event logs used in the evaluation.}
	\label{table-event-logs}
	\resizebox{\columnwidth}{!}{%
\begin{tabular}{l r r r}
	\toprule
	\textbf{Origin} & \textbf{Event log}  & \textbf{Related work}                           &  \\ \midrule
	Synthetic & ConditionalMove & ProDrift  2.0~\cite{DBLP:conf/er/OstovarMRHD16} & \\
	Synthetic & ConditionalRemoval & ProDrift  2.0~\cite{DBLP:conf/er/OstovarMRHD16} & \\
	Synthetic  & ConditionalToSequence  & ProDrift  2.0~\cite{DBLP:conf/er/OstovarMRHD16} & \\
	Synthetic  & Loop & ProDrift  2.0~\cite{DBLP:conf/er/OstovarMRHD16} & \\ 
	Real-world  & Italian help desk\footnotemark[1] & Process~Trees~\cite{quteprints121158}           &  \\
	Real-world  & BPI2011\footnotemark[3]       & ProDrift  2.0~\cite{DBLP:conf/er/OstovarMRHD16} &  \\ \bottomrule
\end{tabular} 

}

\end{wraptable}
We evaluate our approach both on synthetic and real-world event logs.\footnote{\scriptsize \url{https://doi.org/10.4121/uuid:0c60edf1-6f83-4e75-9367-4c63b3e9d5bb}}\footnote{\scriptsize \url{https://doi.org/10.4121/uuid:a7ce5c55-03a7-4583-b855-98b86e1a2b07}}\footnote{\scriptsize \url{https://doi.org/10.4121/uuid:d9769f3d-0ab0-4fb8-803b-0d1120ffcf54} (preprocessed as in~\cite{DBLP:conf/er/OstovarMRHD16})}
We also compare the obtained results with the state-of-the-art methods.
\Cref{table-event-logs} summarizes the event logs used in the evaluation and indicates related work which used these logs.
To discover {\Declare} constraints, we used \gls{minerful}%
\footnote{\scriptsize \url{https://github.com/cdc08x/MINERful}}
because of its high performance~\cite{DBLP:journals/tmis/CiccioM15}.
We opted for the \textit{ruptures} python library%
\footnote{\scriptsize \url{https://github.com/deepcharles/ruptures}}
for change point identification.
We used the \textit{scipy} library%
\footnote{\scriptsize \url{https://docs.scipy.org/doc/scipy/reference/generated/scipy.cluster.hierarchy.linkage.html}} for the clustering of time-series, including the hierarchical clustering.
By experimenting with the clustering algorithm, we tuned the parameters to attain the best outcome, such as the weighted method for linking clusters
(distance between clusters defined as the average between individual points), 
and the correlation metric (to find individual distances between two time-series). 
To enhance {\DriftMap} visualizations, we sort the time-series of each cluster with the mean squared error distance metric.
We implemented our approach in Python 3. Its source code is publicly available.%
\footnote{\scriptsize \url{https://github.com/yesanton/Process-Drift-Visualization-With-Declare}}

\subsection{Detecting drifts}

To demonstrate the accuracy with which our technique detects drifts, we first test it on synthetic data in which drifts were manually inserted, to show that we detect drifts at the points in which they occur. 
Thereafter, we compare our results with the state-of-the-art algorithm ProDrift~\cite{DBLP:conf/er/OstovarMRHD16} 
on real-world event logs.

\subsubsection{Synthetic data.}
\begin{figure}[tb]
	\centering
	\begin{subfigure}[t]{0.31\textwidth}
		\centering
		\includegraphics[width=1\linewidth]{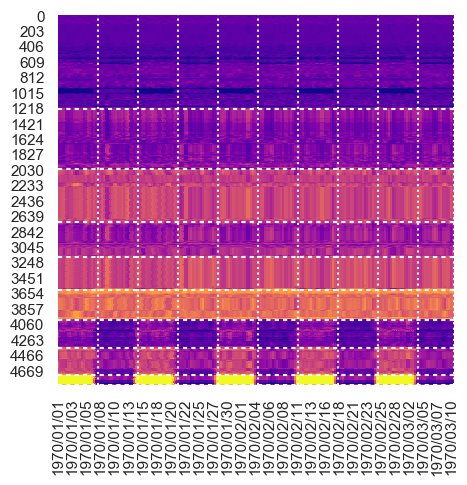}
		\caption{ConditionalMove}
		\label{fig:conditionalMove}
	\end{subfigure}%
	\begin{subfigure}[t]{0.31\textwidth}
		\centering
		\includegraphics[width=1\linewidth]{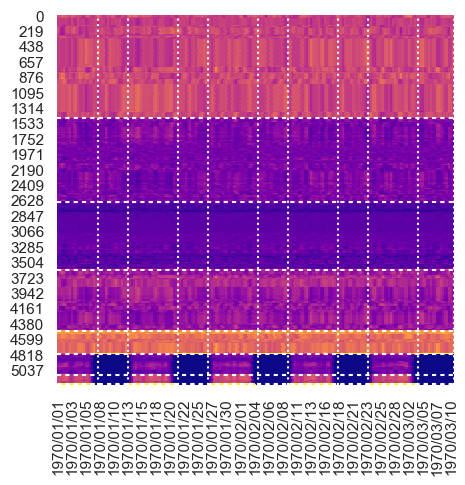}
		\caption{ConditionalRemoval}
		\label{fig:conditionalRemoval}
	\end{subfigure}
	\begin{subfigure}[t]{0.36\textwidth}
		\centering
		\includegraphics[width=1\linewidth]{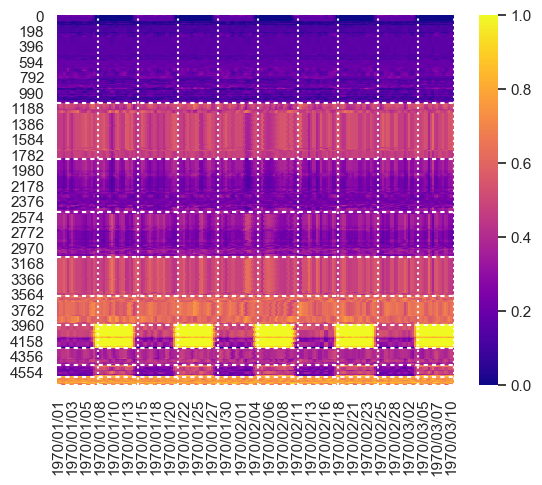}
		\caption{ConditionalToSequence}
		\label{fig:conditionalToSequence}
	\end{subfigure}
	
	
	\begin{subfigure}[t]{0.31\textwidth}
		\centering
		\includegraphics[width=1\linewidth]{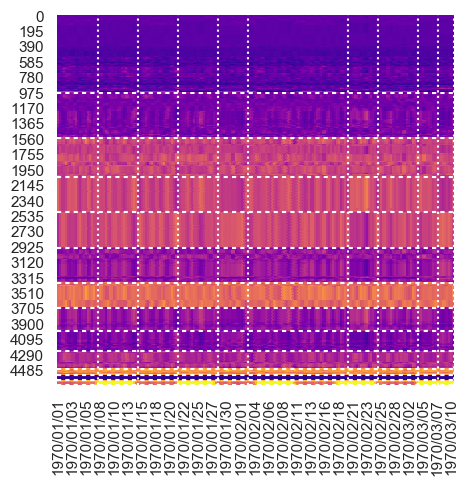}
		\caption{Loop}
		\label{fig:loop}
	\end{subfigure}%
	\begin{subfigure}[t]{0.36\textwidth}
		\centering
		\includegraphics[width=1\linewidth]{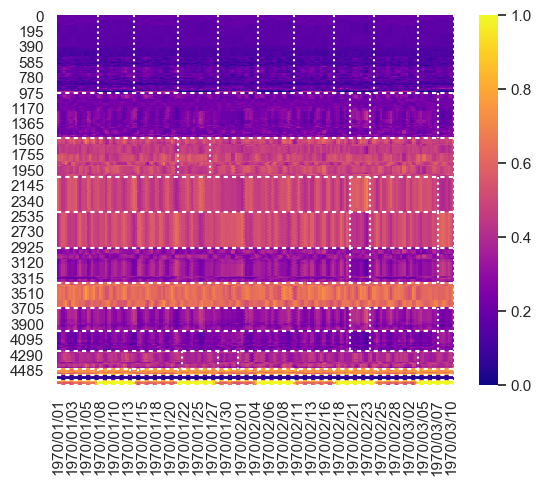}
		\caption{Loop, drifts by cluster}
		\label{fig:loop-in-cluster}
	\end{subfigure}
	\begin{subfigure}[t]{0.31\textwidth}
		\centering
		\includegraphics[width=1\linewidth]{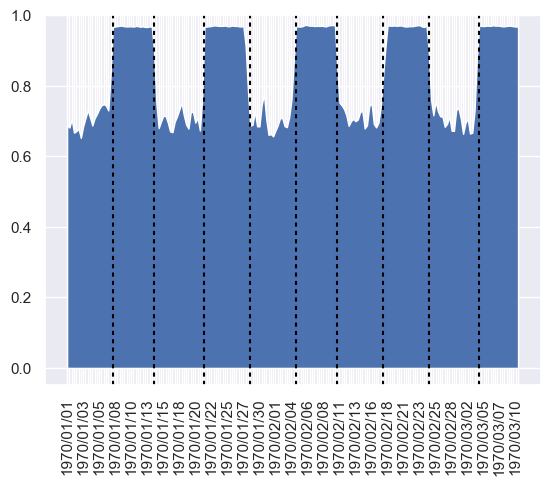}
		\caption{Loop, most erratic cluster}
		\label{fig:loop-detail}
	\end{subfigure}
	\vspace{-1mm}
	\caption{Evaluation results on synthetic logs.}
	\label{fig:manmade}
\end{figure}

Ostovar et al.~\cite{DBLP:conf/er/OstovarMRHD16} published a set of synthetic logs that they altered to artificially include drifting behavior: {ConditionalMove}, {ConditionalRemoval}, {ConditionalToSequence}, and {Loop}.%
\footnote{\scriptsize \url{http://apromore.org/platform/tools}}
\todo{Here, and for ever: please do not use ``textit'' or ``emph'' at random.}
\Cref{fig:manmade} illustrates the results of the application of the \gls{dvd} technique on these logs.
By measuring \emph{precision} as the fraction of correctly identified drifts over all the ones retrieved by \gls{dvd} and \emph{recall} as the fraction of correctly identified drifts over the actual ones, we computed the F-score (harmonic mean of precision and recall) of our results for each log.
Using the default settings and no constraint set clustering, we achieve the F-score of \num{1.0} for logs {ConditionalMove}, {ConditionalRemoval}, {ConditionalToSequence}, and \num{0.89} for the {Loop} log. 
When applying the cluster-based change detection for the {Loop} log, we achieve the F-score of \num{1.0}. 
The {\DriftMap} for the \emph{Loop} log is depicted in~\cref{fig:loop-in-cluster}. In contrast to~\cite{DBLP:conf/er/OstovarMRHD16} we can see which behavior in which cluster contributes to the drift. The {\DriftChart} in~\cref{fig:loop-detail} illustrates the trend of confidence for the most erratic cluster for the \emph{Loop} log.

%

\subsubsection{Real-world data.}
\label{italian-ticket-log-eval}
%
\begin{figure}[tb]
	\centering
	\begin{subfigure}[t]{0.315\textwidth}
		\centering
		\includegraphics[width=1\linewidth]{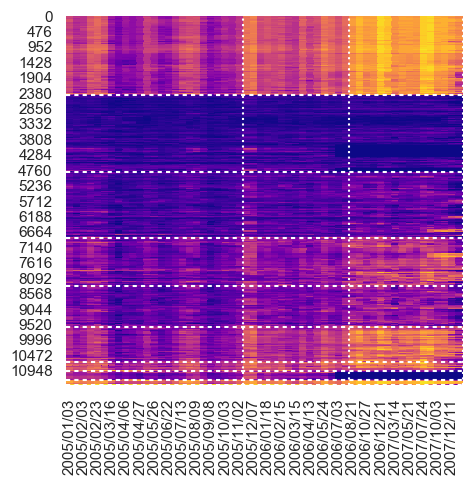}
		\caption{Overall change points}
		\label{fig:bpic2011:alldrifts}
	\end{subfigure}%
	\begin{subfigure}[t]{0.35\textwidth}
		\centering
		\includegraphics[width=1\linewidth]{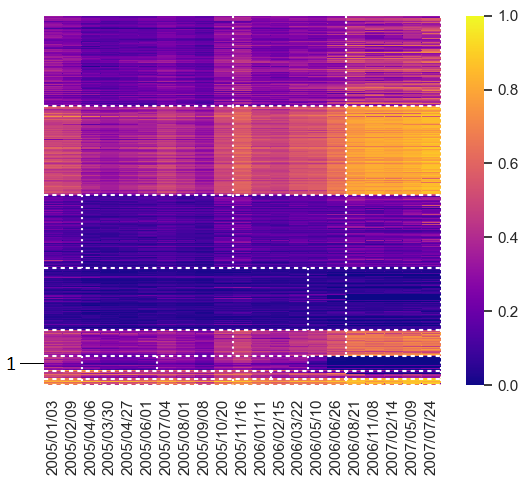}
		\caption{Drifts by cluster}
		\label{fig:bpic2011:separate}
	\end{subfigure}%
	\begin{subfigure}[t]{0.33\textwidth}
		\centering
		\includegraphics[width=1\linewidth]{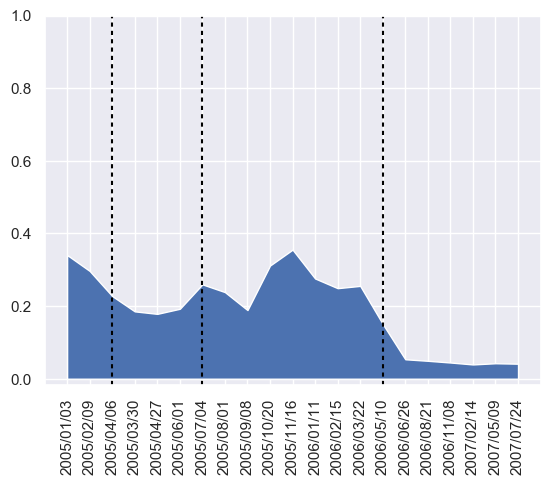}
		\caption{Most erratic cluster}
		\label{fig:bpic2011:erratic}
	\end{subfigure}
	
	\caption{BPIC2011 hospital log \gls{dvd} visualizations.}
	\label{fig:bpic2011}
	\vspace{-2mm}
\end{figure}
\Cref{fig:bpic2011:alldrifts} illustrates the {\DriftMap} constructed for the {BPIC2011 event log}.\footnotemark[3]
As in~\cite{DBLP:conf/er/OstovarMRHD16}, two drifts are detected towards the second half of the time span of the log.
However, in addition, our technique identifies drifting behavior at a finer-granular level.
\Cref{fig:bpic2011:separate} shows the drifts pertaining to clusters of constraints.
The trend of the confidence measure for the most erratic cluster is depicted in~\cref{fig:bpic2011:erratic}.

\begin{figure}[tb]
	\centering
	\begin{subfigure}[t]{0.297\textwidth}
		\centering
		\includegraphics[width=1\linewidth]{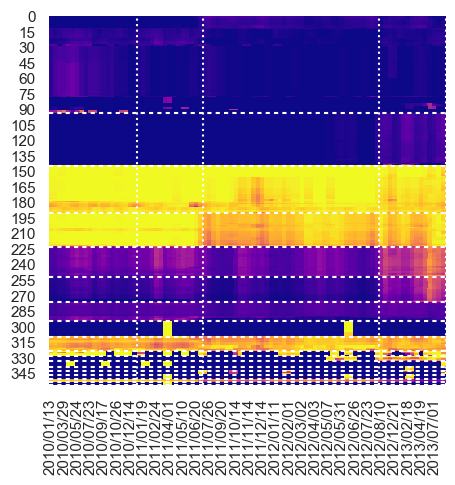}
		\caption{Overall change points}
		\label{fig:italianhelp:alldrifts}
	\end{subfigure}%
	\begin{subfigure}[t]{0.36\textwidth}
		\centering
		\includegraphics[width=1\linewidth]{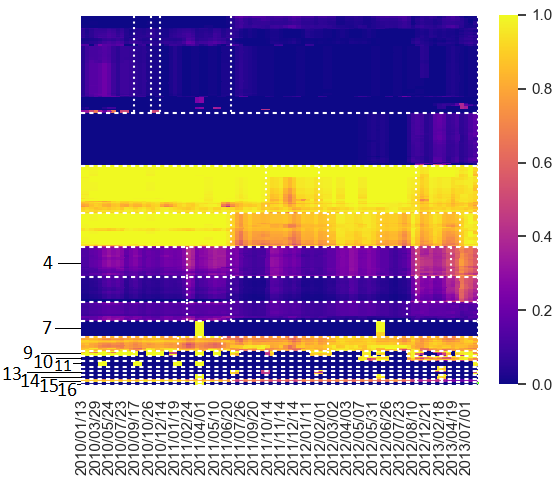}
		\caption{Drifts by cluster}
		\label{fig:italianhelp:separate}
	\end{subfigure}
	\begin{subfigure}[t]{0.33\textwidth}
		\centering
		\includegraphics[width=1\linewidth]{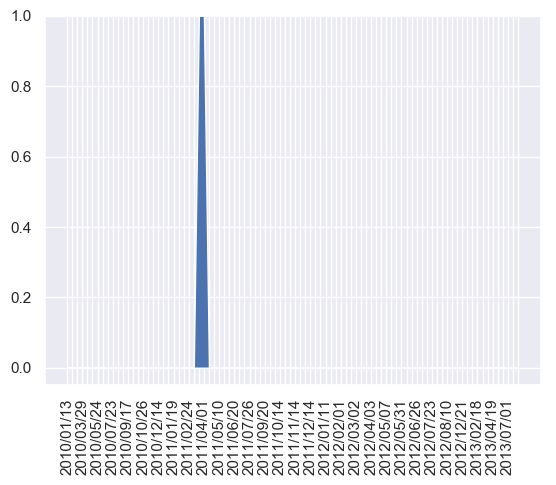}
		\caption{Most erratic cluster}
		\label{fig:italianhelp:outlier}
	\end{subfigure}
	
	\caption{Italian help desk log \gls{dvd} visualizations.}
	\label{fig:italianhelp}
	\vspace{-2mm}
\end{figure}

Our technique correctly detects drifts in the {Italian help desk} log, by identifying the same two drifts that were found by ProDrift~\cite{quteprints121158}, approximately in the first half and towards the end of the time span.
As illustrated  by the \gls{dvd} visualization in~\cref{fig:italianhelp:alldrifts}, in addition we detected another sudden change in the first quarter.
Following on that, we analyzed the within-cluster changes (\cref{fig:italianhelp:separate}) and noticed that the most erratic cluster contains an outlier, as is shown by the spike in~\cref{fig:italianhelp:outlier}.

\subsection{Explaining drifts}
\label{section:explaining-drifts}
\begin{figure}[tb]
	\centering
	\begin{subfigure}{0.33\textwidth}
		\centering
		\includegraphics[width=1\linewidth]{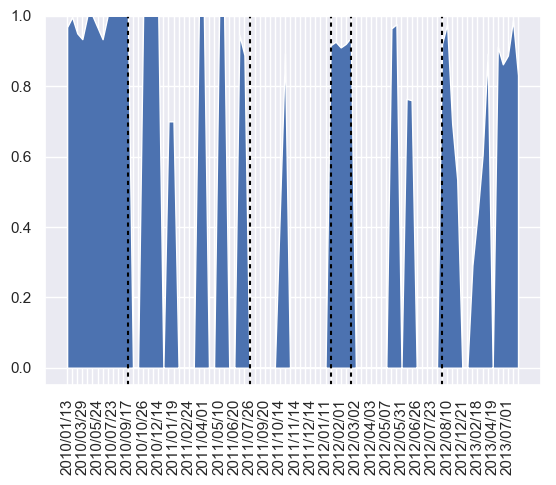}
		\caption{Cluster 9, \Errtcsm: 780.04}
		\label{fig:italianticket:9}
	\end{subfigure}%
	\begin{subfigure}{0.33\textwidth}
		\centering
		\includegraphics[width=1\linewidth]{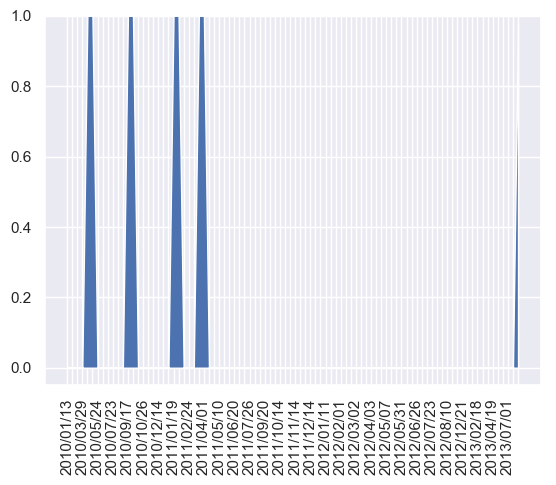}
		\caption{Cluster 11, \Errtcsm: 328.88}
		\label{fig:italianticket:11}
	\end{subfigure}
	\begin{subfigure}{0.33\textwidth}
		\centering
		\includegraphics[width=1\linewidth]{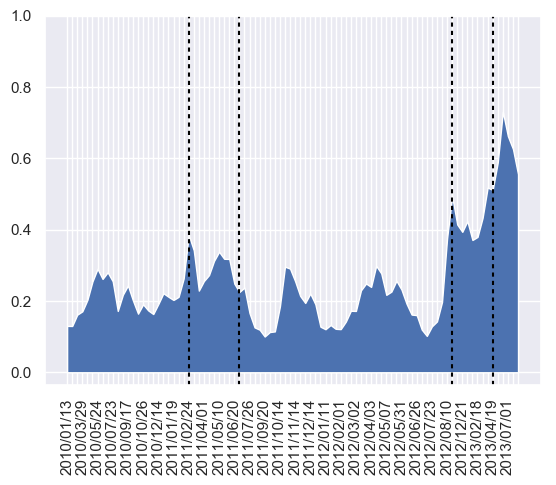}
		\caption{Cluster 4, \Errtcsm: 196.01}
		\label{fig:italianticket:4}
	\end{subfigure}
	\caption{Italian help desk log detailed clusters.}
	\label{fig:italian_ticket_detailed}
\end{figure}

%
\begin{wraptable}[15]{r}{.33\textwidth}
	\caption{Italian help desk log erratic clusters.}
	\label{table:italianticekt-err-clusters}
\rowcolors{2}{white}{gray!12.5}
\begin{tabular}{r D{3}}
	\toprule
	\textbf{Drift number} & \textbf{{\Errtcsm} measure} \\ 
	
	\midrule
    without drift & 89\\
    \textbf{9} & 780.041 \\
    \textbf{11} & 328.881 \\
    \textit{14} & 293.887 \\
    \textit{10} & 292.712 \\
    \textit{13} & 289.103 \\
    \textit{7} & 232.401 \\
    \textbf{4} & 196.012 \\
    \textit{15} & 171.012 \\
    \textit{16} & 166.111 \\
    
	\bottomrule
\end{tabular} 


\end{wraptable}
To better understand a particular drift, we further examine the constraints that participate in the drift. 
Using the example of the {Italian help desk log} presented above, we examine the most erratic behavior clusters' drifts (calculated using~\cref{eqn:erratic}), shown in~\cref{table:italianticekt-err-clusters}. 
\begin{table}[bt]
	\resizebox{0.9\columnwidth}{!}{
		\resizebox{\columnwidth}{!}{%
\rowcolors{2}{white}{gray!12.5}
\begin{tabular}{r r r r D{1} D{1} D{1}}
	\toprule
	                         \textbf{Cluster} & \textbf{Constraint} & \textbf{Activity 1}         & \textbf{Activity 2}         & \textbf{Min} & \textbf{Max} & \textbf{Mean} \\ \midrule
	                        \cellcolor{white} & {\ChaPrecTmp}       & \Task{Take in charge ticket} & \Task{Create SW anomaly}     & 0.0          & 100          & 42.8          \\
	                      \multirow{-2}{*}{9} & {\AltPrecTmp}       & \Task{Assign seriousness}    & \Task{Create SW anomaly}     & 0.0          & 100          & 49.0          \\ \midrule
	                    \cellcolor{gray!12.5} & {\ChaPrecTmp}       & \Task{Take in charge ticket} & \Task{Schedule intervention} & 0.0          & 100          & 9.9           \\
	\multirow{-2}{*}{\cellcolor{gray!12.5}11} & {\AltPrecTmp}       & \Task{Assign seriousness}    & \Task{Schedule intervention} & 0.0          & 100          & 9.9           \\ \midrule
	                        \cellcolor{white} & {\ChaRespTmp}       & \Task{Take in charge ticket} & \Task{Wait}                  & 9.4          & 69.6         & 23.2          \\
	                        \cellcolor{white} & {\NotSuccTmp}       & \Task{Resolve ticket}        & \Task{Wait}                  & 10           & 77.2         & 26            \\
	                        \cellcolor{white} & {\NotSuccTmp}       & \Task{Wait}                  & \Task{Assign seriousness}    & 10           & 78           & 26.6          \\
	                        \cellcolor{white} & {\NotSuccTmp}       & \Task{Wait}                  & \Task{Take in charge ticket} & 9.8          & 73.3         & 22.1          \\
	                        \cellcolor{white} & {\AltRespTmp}       & \Task{Assign seriousness}    & \Task{Wait}                  & 9            & 72.3         & 23.8          \\
	                        \cellcolor{white} & {\AltRespTmp}       & \Task{Wait}                  & \Task{Closed}                & 8.3          & 61.4         & 22.5          \\
	                        \cellcolor{white} & {\AltRespTmp}       & \Task{Wait}                  & \Task{Resolve ticket}        & 8.3          & 61.4         & 22.8          \\
	                      \multirow{-8}{*}{4} & {\UniqTmp}          & \Task{Wait}                  &                              & 9.8          & 68.6         & 25.1          \\ \bottomrule
\end{tabular} 

}

	}
	\caption{Italian ticket log constraints; including min, max, and mean confidence.}
	\label{table:italianticekt-drifts-constrains}
\end{table}
In~\cref{fig:italian_ticket_detailed}, we present the most erratic examples of behavior, and in~\cref{table:italianticekt-drifts-constrains} we present the constraints that describe that specific behavior after applying the constraint minimization algorithm.

\Cref{fig:italianticket:9} shows an erratic behavior, which visually corresponds to the \emph{reoccurring concept} classification from~\cref{fig:driftypes2}. 
Examining the constraints that constitute this behavior, the analyst could conclude that in the dates of the peak in~\cref{fig:italianticket:9} the activity \Task{Create SW anomaly} always had \Task{Take in charge ticket} executed immediately beforehand, and otherwise in the other parts of the plot. 
Also, she could conclude that before \Task{Create SW anomaly} the \Task{Assign seriousness} activity was executed, and no other \Task{Create SW anomaly} occurred in between.

\begin{sloppypar}
	\Cref{fig:italianticket:11} has four spikes, where \Task{Schedule intervention} activities occurred. 
	Immediately before \Task{Schedule intervention}, \Task{Take in charge ticket} occurred.
	Also, \Task{Assign seriousness} occurred had to occur before \Task{Schedule intervention} recurred. 
	We notice, however, that this cluster shows \emph{outlier} behavior, due to its rare changes.
\end{sloppypar}
\begin{sloppypar}
	Finally, \cref{fig:italianticket:4} depicts a \emph{gradual} drift until June 2012, and the \emph{incremental} drift afterward. 
	We notice that all constraints in the cluster have \Task{Wait} either as an activation (e.g., with \AltResp{\Task{Wait}}{\Task{closed}}) or as a target (e.g., with \ChaResp{\Task{Take in charge ticket}}{\Task{Wait}}).
\end{sloppypar}

\vspace{-2mm}
\subsection{Discussion}
\vspace{-1mm}

Our method addresses all the five requirements for process drift detection presented in~\cref{sec:problem} as follows:

\begin{compactitem}
	\item[\ref{req:identify}]
	We evaluated our method with the synthetic logs showing its ability to identify drifts precisely;
	\item[\ref{req:categorize}]
	We developed a visualization approach based on {\DriftMap}s and {\DriftChart}s for classification of process drifts and have shown its effectiveness for real-world logs. Our enhanced approach based on change point detection has yielded effective to automatically discover the exact points at which \textit{sudden} and \textit{reoccurring concept} drifts occur.
	The indicative approximation of long-running progress of \textit{incremental} and \textit{gradual} drifts was also found.
	\emph{Outliers} were detected via time series clustering;
	\item[\ref{req:drill}] 
	Using clustering, {\DriftMap}s, and {\DriftChart}s, the method enables the drilling down into (rolling up out) sections with a specific behavior (general vs. cluster-specific groups of constraints);
	\item[\ref{req:quanti}]
	We introduced, and incorporated into our technique, a drift measure called {\Errtcsm} that quantifies the extent of the drift change;
	\item[\ref{req:quali}]
	To further qualitatively analyze the detected drifts, \gls{dvd} shows how the process specification looks before and after the drift (as a list of {\Declare} constraints, refer to~\Cref{table:italianticekt-drifts-constrains}).
\end{compactitem}

\noindent
We found that the size of the window does not introduce significant difference in results for the automatic evaluation of the \gls{dvd}, so we recommend using the number of windows that will guide the visual search best, that is around 60 windows should be produced for one graph. That means the recommended parameters are: $\smash{\WinStep = \frac{|L|}{60 + 1}}$ and $\smash{\WinSize = 2\cdot\WinStep}$ for smooth visual representation.

%
%
\section{Conclusions}
\label{sec:conclusion}

In this paper, we presented a visual technique for detecting and analyzing process drifts in logs of executed business processes. 
First, the technique uses the MINERful technique to discover declarative process constraints from logs. 
Second, it applies clustering and change point detection methods over time series of characteristics of the discovered constraints to detect process drifts (in parts of) business processes. 
The technique then devises visualizations of the detected clusters and change points for the visual classification of drifts. 
Finally, we presented a technique for evaluating and explaining process drifts.

We evaluated our technique both on synthetic and real-world data. 
On synthetic logs, the technique achieved an average F-score of $0.96$ and outperformed all the state-of-the-art methods. 
On real-world logs, the technique managed to describe all types of process drifts in a comprehensive manner. 
Also, the evaluation reported that our technique can identify outliers of process behavior.

Limitations of the work at hand naturally give rise to future research.
First, one can study the problem of automatic classification of process drifts; we plan to use shapelets~\cite{DBLP:journals/datamine/AbandaML19} to solve this problem.
Second, one can study how the use of other declarative process constraints, e.g., the 4C spectrum~\cite{DBLP:conf/apn/PolyvyanyyWCRH14} or branched {\Declare}~\cite{DBLP:journals/is/CiccioMM16}, impacts the effectiveness of the technique.
Third, an empirical evaluation with the potential users of the technique can provide further insights for improving the usability of the approach.
Finally, we argue that, based on the identified past process drifts, and using time-series prediction algorithms, one can predict future drifts to prepare for forecasted changes~\cite{DBLP:conf/bpm/PollPRRR18}.

\medskip
\noindent
\textbf{Acknowledgements.}
This work is partially funded by the EU H2020 program under MSCA-RISE agreement 645751 ({RISE\_BPM}).
Artem Polyvyanyy was partly supported by the Australian Research Council Discovery Project DP180102839.

%
%

%
%
%
\end{document}